\documentclass[letterpaper]{article} 
\usepackage{aaai24}  
\usepackage{times}  
\usepackage{helvet}  
\usepackage{courier}  
\usepackage[hyphens]{url}  
\usepackage{graphicx} 
\usepackage[T1]{fontenc}
\usepackage{amsmath}
\usepackage{xurl}
\usepackage{adjustbox} 
\usepackage{booktabs}
\usepackage{array}
\usepackage{natbib}  
\usepackage{caption} 
\usepackage{algorithm}
\usepackage{algorithmic}
\newcommand{\ie}{\textit{i}.\textit{e}., }
\newcommand{\eg}{\textit{e}.\textit{g}., }

\usepackage[usenames,dvipsnames]{xcolor}
\newcommand{\answerYes}[1]{\textcolor{blue}{#1}} 
 
\newcommand{\answerNA}[1]{\textcolor{gray}{#1}} 
 
\usepackage{newfloat}
\usepackage{listings}
\DeclareCaptionStyle{ruled}{labelfont=normalfont,labelsep=colon,strut=off} 
\floatstyle{ruled}
\newfloat{listing}{tb}{lst}{}
\floatname{listing}{Listing}
\title{Human vs. LMMs: Exploring the Discrepancy in Emoji Interpretation and Usage in Digital Communication}
\author{
    Hanjia Lyu\equalcontrib\textsuperscript{\rm 1},
    Weihong Qi\equalcontrib\textsuperscript{\rm 1},
    Zhongyu Wei\textsuperscript{\rm 2},
    Jiebo Luo\textsuperscript{\rm 1}}
\affiliations {
    \textsuperscript{\rm 1}University of Rochester\\
    \textsuperscript{\rm 2}Fudan University\\
    hlyu5@ur.rochester.edu,
    jluo@cs.rochester.edu
}

\usepackage{bibentry}

\begin{document}

\maketitle

\begin{abstract}
Leveraging Large Multimodal Models (LMMs) to simulate human behaviors when processing multimodal information, especially in the context of social media, has garnered immense interest due to its broad potential and far-reaching implications. Emojis, as one of the most unique aspects of digital communication, are pivotal in enriching and often clarifying the emotional and tonal dimensions. Yet, there is a notable gap in understanding how these advanced models, such as GPT-4V, interpret and employ emojis in the nuanced context of online interaction. This study intends to bridge this gap by examining the behavior of GPT-4V in replicating human-like use of emojis. The findings reveal a discernible discrepancy between human and GPT-4V behaviors, likely due to the subjective nature of human interpretation and the limitations of GPT-4V's English-centric training, suggesting cultural biases and inadequate representation of non-English cultures.
\end{abstract}

\section{Introduction}

The advent of Large Multimodal Models (LMMs) has marked a significant milestone in the use of machine intelligence to simulate human behaviors while perceiving multimodal information~\cite{park2023generative, fui2023generative}. This field of research, particularly when applied to social media, has attracted considerable attention due to its vast potential and implications~\cite{törnberg2023simulating,gao2023s3}. One of the most unique aspects of digital communication is the use of emojis, which, in this context, are not mere embellishments but fundamental components that enhance and often clarify the emotional and tonal aspects of communication~\cite{incontext,intention}. 

While extensive research has been dedicated to the domain of the image-text pair understanding capabilities of LMMs~\cite{yang2023dawn,lyu2023gpt,zhang2024cocot}, there remains a notable gap in comprehending how these sophisticated models navigate the nuanced landscape of emoji-enhanced communication. The focus of this study lies in understanding whether LMMs can accurately emulate how humans employ emojis in digital communication. GPT-4V(ision), as one of the most advanced large multimodal models, represents the cutting edge in AI development~\cite{gpt4report,fu2023mme,yu2023mm}. Its capabilities in processing multimodal content make it an ideal candidate for studying complex interpretative tasks like emoji understanding. Our study focuses on two research questions:
\begin{itemize}
    \item \textbf{RQ1:} How does {\sc GPT-4V}'s interpretation of emojis compare with that of humans? 
    \item \textbf{RQ2:} Does {\sc GPT-4V} employ emojis in writing social media posts in a manner that differs from human usage?
\end{itemize}

To investigate these two questions, we conduct two studies. First, we compare the semantic interpretations that {\sc GPT-4V} associates with emojis against those attributed by humans. Second, to obtain more in-depth insights, we prompt {\sc GPT-4V} to generate social media posts incorporating emojis. The emojis selected by {\sc GPT-4V} are then compared with the emojis used by humans in similar scenarios. 

This research strives to enhance the collective understanding of AI's strengths and limitations in decoding and using modern symbolic language. The insights gleaned here are intended to inform the ongoing development of AI systems that are not only technologically advanced but also embedded with a deeper sense of empathy and cultural sensitivity.

\section{Related Work}\label{sec:related_work}

A substantial body of literature focuses on the understanding of emoji use in social media, contributing to the development of sophisticated analytical tools for emojis prior to the advent of foundation models. For instance, \citet{Kralj2015emojis} introduced a sentiment lexicon specifically tailored for emojis, reflecting their practical use in social media. \citet{eisner2016emoji2vec} and \citet{liu2021improving} developed specialized embeddings for emojis. Complementing these studies, \citet{rodrigues2018lisbon} created a dataset that codifies the norms of emoji use across seven distinctive dimensions.

Since the emergence of Large Language Models (LLMs), several studies have explored the capability of LLMs like ChatGPT in comprehending emojis to enhance sentiment analysis~\cite{kocon2023chatgpt, kheiri2023sentimentgpt}. These investigations primarily input emojis as Unicode characters. However, this approach has notable limitations. The rich visual nuances present in emoji images are often not fully captured in their Unicode form. Moreover, using image-based emojis could mirror more closely how humans perceive mixed media. The visual aspect of emojis is particularly crucial, considering their inherent ambiguity and the diverse interpretations they may elicit, which can lead to significant misinterpretations in social media interactions~\cite{incontext, contextfree}.

\section{Experiments}\label{sec:exp}
In this section, we detail the experimental setup and results of the two studies we conduct. Study 1 concentrates on examining {\sc GPT-4V}'s interpretation of emojis, while Study 2 investigates its usage of emojis. For these experiments, we employ the {\tt gpt-4-vision-preview} variant, which was the latest as of December 2023. It is important to note that, in our prompts, emojis are inputted as \textit{images}.

\subsection{Study 1: Emoji Interpretation}\label{sec:study1_interpretation}

\subsubsection{Study Design}
To investigate RQ1, which assesses whether or not {\sc GPT-4V} interprets emojis differently from humans, we prompt {\sc GPT-4V} to describe each emoji using a single word. The generated word is then compared with the word chosen by humans to describe the same emoji. 

\subsubsection{Emoji Selection}\label{sec:study1_method} We use the emojis from the dataset compiled by \citet{contextfree}. This dataset encompasses a carefully curated collection of 1,289 commonly-used emojis. Each emoji in this dataset has been categorized into one of the 20 fine-grained groups, including but not limited to categories like \textit{objects}, \textit{nature}, and \textit{travel places}. Additional detail can be found in \citet{contextfree}.

\subsubsection{Prompting GPT-4V} Motivated by \citet{contextfree}, to understand how {\sc GPT-4V} interprets each emoji, we present the emoji to {\sc GPT-4V} in an image form, accompanied by the instruction: ``\textit{Describe the emoji with a single, accurate word}''. We maintain the default {\tt temperature} setting at a value of 1. Each emoji-prompt pair is inputted into {\sc GPT-4V} multiple times, in order to compile a comprehensive vocabulary dictionary from {\sc GPT-4V}’s responses for each emoji. Specifically, we repeat this process six times for every pair. The number is selected based on a pilot study where we find that the responses remain relatively stable after repeating the process six times.

\subsubsection{Human Annotations} We use the human annotations collected by \citet{contextfree}. Here we briefly discuss their annotation process. They recruited participants via Amazon Mechanical Turk (AMT). Participants, who were required to be English-speaking U.S. residents over 18 with a high approval rate and experience on AMT, were tasked with describing emojis using a single word. 

\begin{figure}[t]
    \centering
    \includegraphics[width=\linewidth]{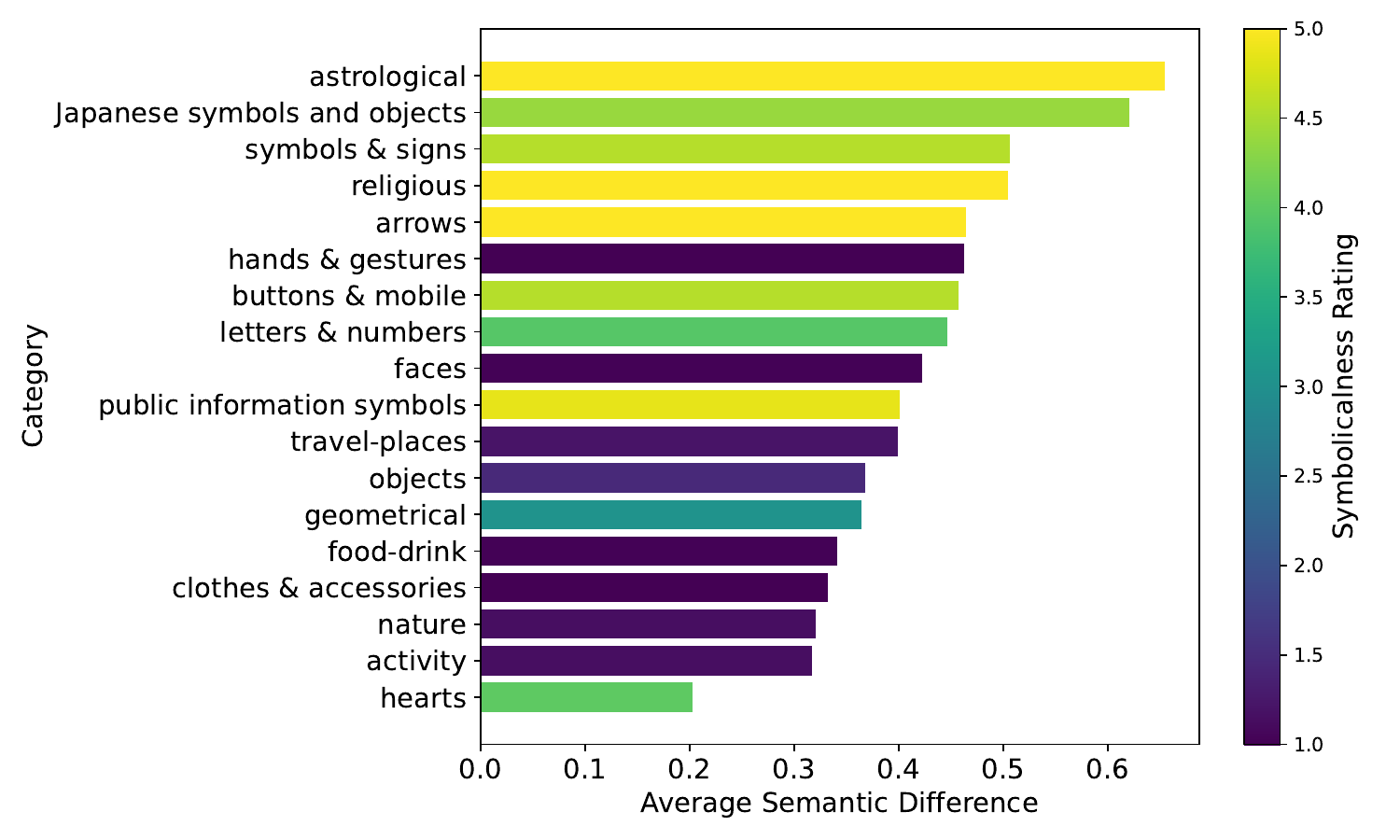}
    \caption{Variations in semantic interpretation between {\sc GPT-4V} and humans for emojis across different categories.}
    \label{fig:semantic_difference}
\end{figure}

\subsubsection{Results} Instead of directly comparing the words that humans and {\sc GPT-4V} use to describe emojis, a more meaningful approach is to compare whether the words convey different semantic meanings. Following the methodology of \citet{contextfree}, we transform each word into its vector representation using GloVe vectors~\cite{pennington2014glove}. We then measure the semantic dispersion between humans and {\sc GPT-4V} for the same emoji by calculating the centroid distance between two clusters: one comprising the word embeddings of words used by humans, and the other of those generated by {\sc GPT-4V}. For this computation, we employ cosine distance.

Figure~\ref{fig:semantic_difference} shows that \textbf{there is a varying level of differences in interpretation between GPT-4V and humans}. Notably, the greatest differences are seen in categories like \textit{astrological} (\eg \includegraphics[height=1em]{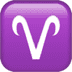}, \includegraphics[height=1em]{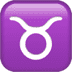}, \includegraphics[height=1em]{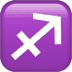}), \textit{Japanese symbols and objects} (\eg \includegraphics[height=1em]{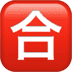}, \includegraphics[height=1em]{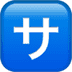}, \includegraphics[height=1em]{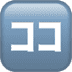}), and \textit{religious} (\eg \includegraphics[height=1em]{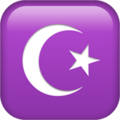}, \includegraphics[height=1em]{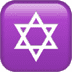}, \includegraphics[height=1em]{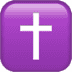}). In contrast, categories like \textit{hearts} (\eg \includegraphics[height=1em]{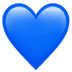}, \includegraphics[height=1em]{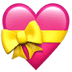}, \includegraphics[height=1em]{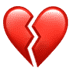}), \textit{activity} (\eg \includegraphics[height=1em]{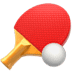}, \includegraphics[height=1em]{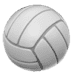}, \includegraphics[height=1em]{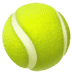}), and \textit{nature} (\eg \includegraphics[height=1em]{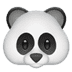}, \includegraphics[height=1em]{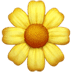}, \includegraphics[height=1em]{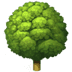}) exhibit the lowest levels of discrepancies. The bar colors in Figure~\ref{fig:semantic_difference} represent the symbolicalness ratings for each category, quantifying the extent to which an emoji is perceived as part of a conventional set of symbols. The symbolicalness of each category was annotated by \citet{contextfree}. It is evident that the interpretative divergence between {\sc GPT-4V} and humans is more pronounced in categories where emojis are predominantly viewed as symbolic.

The observed discrepancy in emoji interpretation between {\sc GPT-4V} and humans likely stems from the level of \textbf{ambiguity} in emoji interpretations. Symbolic emojis often embody abstract concepts that are subject to individual interpretation. Humans, drawing on their unique experiences, beliefs, and perceptions, bring a highly subjective understanding to these symbols. In contrast, {\sc GPT-4V}'s interpretations are constrained by its training dataset. If this dataset lacks diversity, particularly in the context of symbolic emojis used across various cultural backgrounds, the model's capacity for accurate interpretation is notably diminished. To deepen our understanding of these interpretative variances, we then quantitatively assess the level of ambiguity exhibited by both {\sc GPT-4V} and humans in their emoji interpretation.

\begin{figure}[t]
    \centering
    \includegraphics[width=\linewidth]{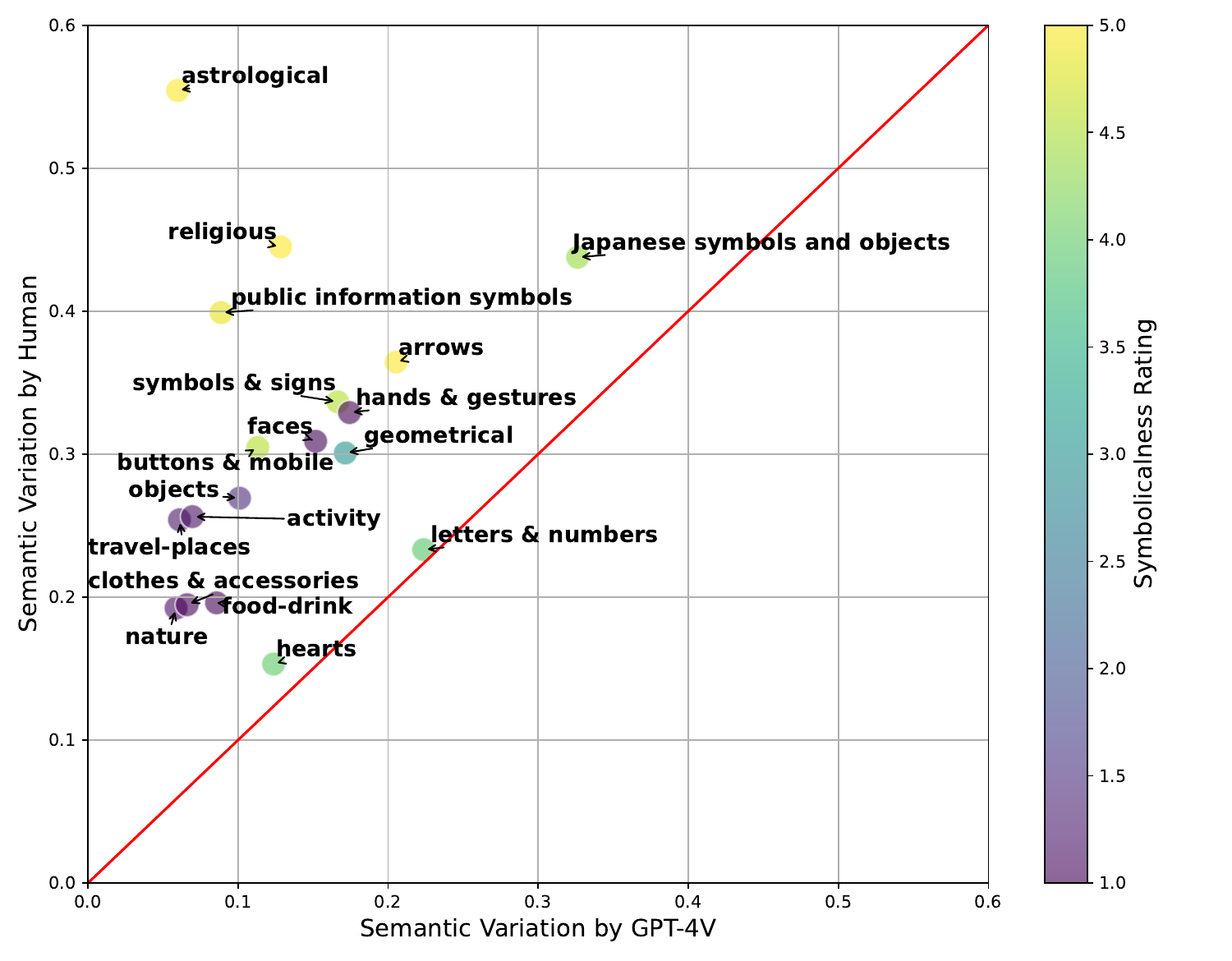}
    \caption{Comparison of the degree of ambiguity in emoji interpretation between humans and {\sc GPT-4V} across different emoji categories, along with the symbolicalness rating of these categories. The red diagonal line represents the point of equality where {\sc GPT-4V} and humans exhibit the same level of ambiguity in emoji interpretation.}
    \label{fig:ambiguity}
\end{figure}

\subsubsection{Ambiguity Analysis} For a given emoji $e$, let $\mathcal{V}$ be the set of unique words used for its annotations by {\sc GPT-4V}. We denote the most commonly used word as $v^{*}$ and its embedding as $\varepsilon_{v^{*}}$. The embedding for any other word $v$ in $\mathcal{V}$ is represented as $\varepsilon_{v}$. To quantify the semantic variation $\mathcal{SV}_{e}$ for emoji $e$, we compute the weighted sum of cosine distance between $\varepsilon_{v^{*}}$ and $\varepsilon_{v}$, as follows:
\begin{equation}
\mathcal{SV}_{e} = \sum_{v \in \mathcal{V}} f_{v} \cdot (1 - \text{cosine\_similarity}(\varepsilon_{v}, \varepsilon_{v^{*}}))
\end{equation}
where $f_{v}$ denotes the frequency of word $v$. The ambiguity level in emoji interpretations by humans is collected from \citet{contextfree}.

Figure~\ref{fig:ambiguity} presents a comparative analysis of the ambiguity in emoji interpretations between humans and {\sc GPT-4V} across various emoji categories. Notably, \textbf{GPT-4V demonstrates less interpretative ambiguity compared to humans}, consistent with the findings from \citet{contextfree}. This difference can be attributed to the varying levels of background knowledge; humans lacking specific knowledge might inaccurately describe symbolic emojis, leading to a wider array of descriptive words. {\sc GPT-4V}, with its extensive training, is less prone to such variance.

The red diagonal line in the figure marks the equality in ambiguity levels between {\sc GPT-4V} and humans. Data points, colored warmer (\ie more yellow) as they diverge from this line, indicate a correlation between the symbolic nature of emojis and the disparity in interpretative ambiguity between {\sc GPT-4V} and humans.

Interestingly, an outlier is observed in {\sc GPT-4V}'s interpretation ambiguity, particularly with \textit{Japanese symbols and objects}. Upon analysis, we find that it is caused by {\sc GPT-4V}'s frequent misinterpretations in this category. We hypothesize that the \textbf{misinterpretations result from the limitations in GPT-4V's training data which is primarily sourced from English corpora}. This limitation likely hinders its understanding of culturally specific emojis. Similarly, the higher ambiguity in human annotations for this category can be attributed to the predominantly English-speaking U.S. resident demographic of the participants in \citet{contextfree}'s study, suggesting a cultural bias in emoji interpretation.

\subsection{Study 2: Emoji Usage}\label{sec:study2_usage}

\begin{figure}[t]
    \centering
    \includegraphics[width=\linewidth, trim=1.5cm 1cm 2.5cm 2cm, clip]{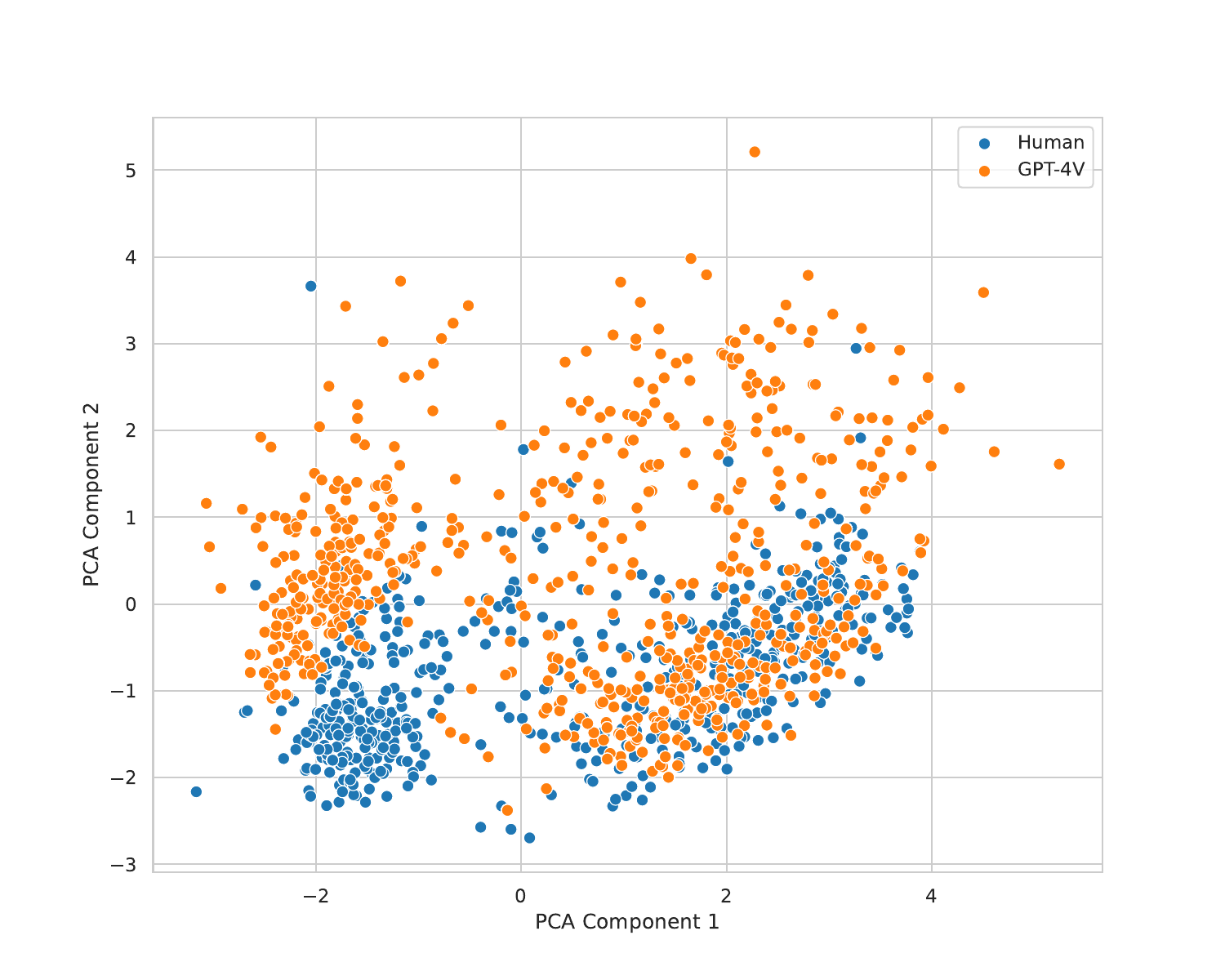}
    \caption{PCA visualization word embeddings of human-used and {\sc GPT-4V} generated emojis.}
    \label{fig: embedding}
\end{figure}

\begin{table}[t]
\centering
\adjustbox{max width=\linewidth}{
\begin{tabular}{{l}c*{2}{c}r}
\toprule[1.1pt]
Topic   & Mean & Median & Standard Error \\
\midrule
Personal experience & 0.2624 & 0.2550 & 0.0061 \\
Pets & 0.2760 & 0.2750 & 0.0080 \\
Family & 0.2472 & 0.2423 & 0.0063 \\
Music & 0.2776 & 0.2754 & 0.0058 \\
Sports & 0.2955 & 0.2904 & 0.0057 \\
\bottomrule[1.1pt]
\end{tabular}}
\caption{The mean and median cosine distances between the emojis used by humans and {\sc GPT-4V}.} 
\label{tab:emb_dist} 
\end{table}

\begin{table*}[t]
\centering
\begin{tabular}{p{3cm}>{\arraybackslash}p{7cm}>{\centering\arraybackslash}p{3cm}>{\centering\arraybackslash}p{3cm}}
\toprule[1.1pt]
Topic  & Text  & Human User & GPT-4V \\
\midrule
Personal experience & October was a fantastic month. Looking forward to more adventures in November and December.  &\includegraphics[height=1.2em]{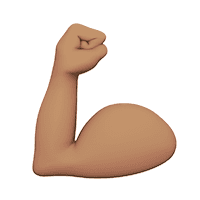}\includegraphics[height=1.2em]{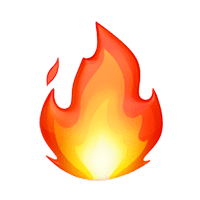}\includegraphics[height=1.2em]{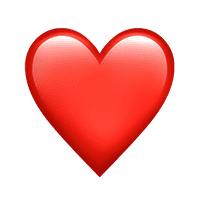}\includegraphics[height=1.2em]{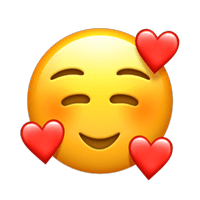}\includegraphics[height=1.2em]{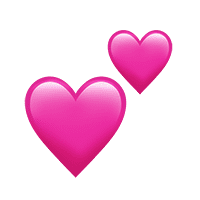}  & \includegraphics[height=1em]{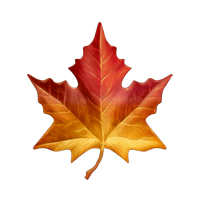}\includegraphics[height=1.2em]{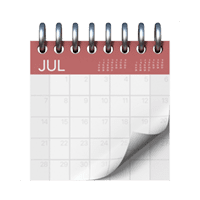}\includegraphics[height=1.2em]{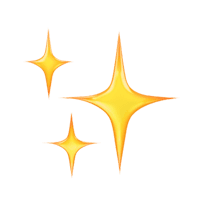} \\
Pets & Enjoying myself and basking in the sunlight!  & \includegraphics[height=1.2em]{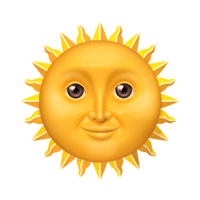}\includegraphics[height=1.2em]{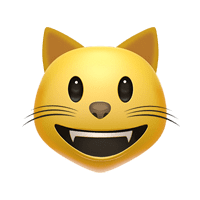}  & \includegraphics[height=1.2em]{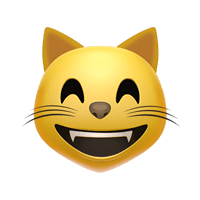}\includegraphics[height=1.2em]{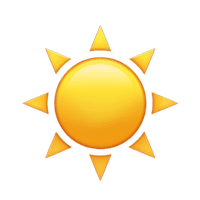}\includegraphics[height=1.2em]{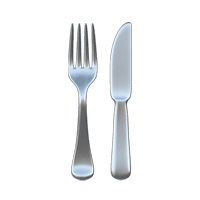} \\
Family & A wedding present for my brother and sister-in-law. Best wishes on their union. & \includegraphics[height=1.2em]{eg_emoji22.png}\includegraphics[height=1.2em]{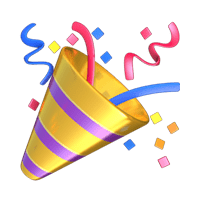} &  \includegraphics[height=1.2em]{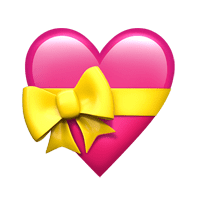}\includegraphics[height=1.2em]{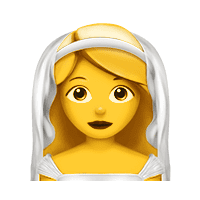}\includegraphics[height=1.2em]{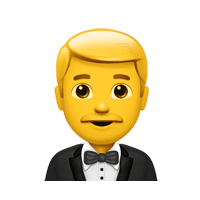}\includegraphics[height=1.2em]{eg_emoji26.png}\\
Music & Step into the booth and I swiftly alter my persona, just like Clark Kent! & \includegraphics[height=1.2em]{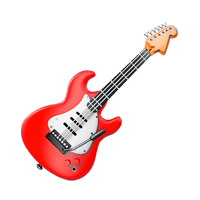}\includegraphics[height=1.2em]{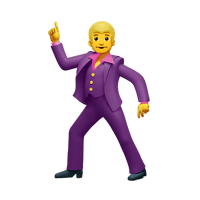} & \includegraphics[height=1.2em]{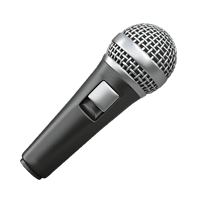}\includegraphics[height=1.2em]{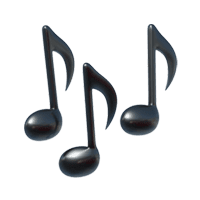}\includegraphics[height=1.2em]{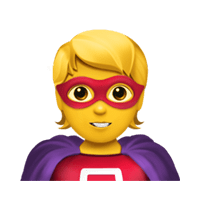} \\
Sports & Extremely happy and proud of Mary, fantastic job! & \includegraphics[height=1.2em]{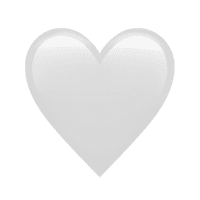}\includegraphics[height=1.2em]{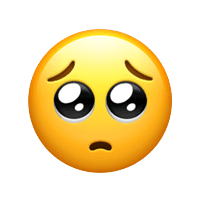}\includegraphics[height=1.2em]{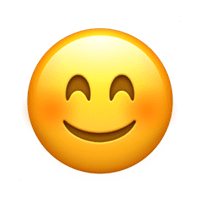}  & \includegraphics[height=1.2em]{eg_emoji32.png}\includegraphics[height=1.2em]{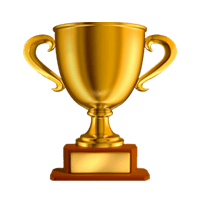}\includegraphics[height=1.2em]{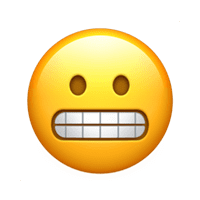} \\
\bottomrule[1.1pt]
\end{tabular}
\caption {Examples of emoji usage of human users and {\sc GPT-4V}. Text has been paraphrased to protect user privacy.} 
\label{tab: emoji_example} 
\end{table*}

\subsubsection{Study Design} While assessing {\sc GPT-4V}'s interpretation of emojis in social media posts is valuable and intriguing, a more straightforward approach involves employing {\sc GPT-4V}'s generative capabilities with social media content and contrasting its emoji usage with that of human users (RQ2). 

\subsubsection{Prompting GPT-4V} We use a specific prompt to elicit the emojis that {\sc GPT-4V} would recommend for a social media post. The prompt is: ``\textit{Imagine you are a social media user seeking the most suitable emojis for your social media posts based on the {\tt context}. Please respond with only three emojis that would be optimal for this purpose}''. We opt to prompt {\sc GPT-4V} to generate three emojis in our study.
This decision is based on our analysis of the social media post dataset we collect, where we find that the average number of emojis per post is 2.50. After filtering out outliers with more than 10 emojis, this average decreases to 2.14.
We will discuss the data collection details in the following sections.

\subsubsection{Context} In this analysis, we examine how {\sc GPT-4V} employs emojis based on given contexts. To simulate a real-world environment, we collect video descriptions from TikTok. We focus on TikTok in the current study because it is one of the most popular social media platforms with a vast, diverse user base. This diversity offers a broad spectrum of real-world data, which is invaluable for understanding emoji usage in different contexts. Additionally, TikTok is known for its trending and up-to-date content. Analyzing emoji usage on this platform can provide insights into current trends and the evolving nature of digital communication. These descriptions, with emojis removed, serve as the context for {\sc GPT-4V}. We then compare the emojis originally used by TikTok users in these descriptions with those suggested by {\sc GPT-4V}. This analysis is designed to investigate the differences in emoji usage between humans and {\sc GPT-4V}, providing insights into the model's understanding and application of emojis in real-world social media contexts.

\subsubsection{Data Collection} Our initial step is to select five prevalent hashtags, indicative of TikTok's most trending topics that span a wide range of themes including sports, pets, music, family, and personal experience. In particular, five hashtags are selected including \#WSL, \#cats, \#family, \#hiphop, and \#bestof2023. Note that WSL stands for the Women's Super League of England. We deliberately choose not to use more generic hashtags like football, Premier League, or NBA in our study. This decision is based on the observation that such hashtags attract a high volume of advertisements, robot-generated content, and unrelated posts, which could skew the authenticity and relevance of the data collected for our analysis. We then employ TikTok's official API to collect videos tagged with these hashtags, specifically those posted from November 1$^{st}$ to December 31$^{st}$, 2023. 
Our dataset includes 5,316 videos and their descriptions, with an average of 255.7 tokens per description. We retain descriptions that contain emojis and exclude those with fewer than three tokens in their non-hashtag text, resulting in 1,689 video descriptions for analysis. The full texts of these video descriptions are inputted into {\sc GPT-4V} as contexts.
To enhance the generalizability of our study, we expand our analysis to include an additional dataset covering 14 new topics, represented by 23 specific hashtags. This supplementary analysis follows the same methodology as the primary study, with the findings detailed in the Appendix for comparison.

\subsubsection{Results}

We direct {\sc GPT-4V} to interpret and summarize the messages conveyed by the emojis used by human users and those generated by {\sc GPT-4V} with a single accurate word. The prompt for the summarization is ``\textit{Describe the message conveyed by the emojis only with one word.}'' To analyze these responses, we use the Bidirectional Encoder Representations from Transformers (BERT) model~\cite{devlin2019bert} to create word embeddings from these one-word descriptions. Next, we calculate the cosine distances between the embeddings derived from human-used emojis and those of {\sc GPT-4V}, thus facilitating a comparative analysis.

 Figure~\ref{fig: embedding} presents the PCA visualization of word embeddings. Generally, the embeddings for {\sc GPT-4V}'s emojis exhibit a sparser distribution compared to those of human-selected emojis, \textbf{indicating a potentially greater variety in emoji selection by GPT-4V}. Notably, there is considerable overlap between the two sets of emojis across all five examined topics, suggesting that \textbf{GPT-4V partially mirrors human emoji usage patterns.} Table~\ref{tab:emb_dist} provides additional insights by quantifying the embedding distances across emoji types. It reveals that the sports topic, represented by \#wsl, has the largest divergence in emoji choices between humans and {\sc GPT-4V}, whereas \#family, representing the family topic, shows the smallest difference. However, it is important to note that the variance in the distances across different topics is not markedly significant, implying that the discrepancies between human and {\sc GPT-4V} emoji usage are relatively consistent regardless of the context. Table~\ref{tab: emoji_example} presents specific examples of emoji usage between human users and {\sc GPT-4V} across the five selected topics.

\section{Discussions and Conclusions}\label{ssec:discussions_conclusions}

In this study, we mainly focus on two research questions to investigate {\sc GPT-4V}'s understanding of emojis, comparing it against human perception. 
This exploration highlights the divergent ways in which emojis are understood by AI and humans, underlined by the subjective nuances of human interpretation and the inherent constraints in {\sc GPT-4V}'s training data. 
Our investigation also reveals {\sc GPT-4V}'s diverse emoji usage in social media contexts, mirroring human patterns to a notable extent. 
These insights are crucial in steering the future development of foundational models, aiming to more accurately replicate nuanced human behaviors online. 
This study, while comprehensive, acknowledges the limitations of focusing on only one large multimodal model whose training corpora are mainly English text. 
Looking forward, we intend to understand training data issues across various LMMs, particularly those trained on non-English corpora. 
Furthermore, the emojis evaluated in this study are predominantly used by English-speaking communities. In our future research endeavors, we plan to broaden the scope of our analysis by including emojis that are commonly used in non-English speaking and diverse cultural communities.

\subsubsection{Code Availability}
Codes are publicly available at \url{https://github.com/VIStA-H/GPT4V-Emoji-Interpretation-Usage}.

\bibliography{aaai24}

\subsection{Paper Checklist}

\begin{enumerate}

\item For most authors...
\begin{enumerate}
    \item  Would answering this research question advance science without violating social contracts, such as violating privacy norms, perpetuating unfair profiling, exacerbating the socio-economic divide, or implying disrespect to societies or cultures?
    \answerYes{Yes.}
  \item Do your main claims in the abstract and introduction accurately reflect the paper's contributions and scope?
    \answerYes{Yes.}
   \item Do you clarify how the proposed methodological approach is appropriate for the claims made? 
    \answerYes{Yes.}
   \item Do you clarify what are possible artifacts in the data used, given population-specific distributions?
    \answerYes{Yes, see the Discussions and Conclusions section.}
  \item Did you describe the limitations of your work?
    \answerYes{Yes, see the Discussions and Conclusions section.}
  \item Did you discuss any potential negative societal impacts of your work?
    \answerYes{Yes, see Appendix.}
      \item Did you discuss any potential misuse of your work?
    \answerYes{Yes, see Appendix.}
    \item Did you describe steps taken to prevent or mitigate potential negative outcomes of the research, such as data and model documentation, data anonymization, responsible release, access control, and the reproducibility of findings?
    \answerYes{Yes, see Appendix.}
  \item Have you read the ethics review guidelines and ensured that your paper conforms to them?
    \answerYes{Yes.}
\end{enumerate}

\item Additionally, if your study involves hypotheses testing...
\begin{enumerate}
  \item Did you clearly state the assumptions underlying all theoretical results?
    \answerNA{NA.}
  \item Have you provided justifications for all theoretical results?
    \answerNA{NA.}
  \item Did you discuss competing hypotheses or theories that might challenge or complement your theoretical results?
    \answerNA{NA.}
  \item Have you considered alternative mechanisms or explanations that might account for the same outcomes observed in your study?
    \answerNA{NA.}
  \item Did you address potential biases or limitations in your theoretical framework?
    \answerNA{NA.}
  \item Have you related your theoretical results to the existing literature in social science?
    \answerNA{NA.}
  \item Did you discuss the implications of your theoretical results for policy, practice, or further research in the social science domain?
   \answerNA{NA.}
\end{enumerate}

\item Additionally, if you are including theoretical proofs...
\begin{enumerate}
  \item Did you state the full set of assumptions of all theoretical results?
    \answerNA{NA.}
	\item Did you include complete proofs of all theoretical results?
    \answerNA{NA.}
\end{enumerate}

\item Additionally, if you ran machine learning experiments...
\begin{enumerate}
  \item Did you include the code, data, and instructions needed to reproduce the main experimental results (either in the supplemental material or as a URL)?
    \answerYes{Yes.}
  \item Did you specify all the training details (e.g., data splits, hyperparameters, how they were chosen)?
    \answerYes{Yes.}
     \item Did you report error bars (e.g., with respect to the random seed after running experiments multiple times)?
    \answerNA{NA.}
	\item Did you include the total amount of compute and the type of resources used (e.g., type of GPUs, internal cluster, or cloud provider)?
   \answerNA{NA.}
     \item Do you justify how the proposed evaluation is sufficient and appropriate to the claims made? 
    \answerYes{Yes.}
     \item Do you discuss what is ``the cost`` of misclassification and fault (in)tolerance?
    \answerNA{NA.}
  
\end{enumerate}

\item Additionally, if you are using existing assets (e.g., code, data, models) or curating/releasing new assets, \textbf{without compromising anonymity}...
\begin{enumerate}
  \item If your work uses existing assets, did you cite the creators?
    \answerYes{Yes.}
  \item Did you mention the license of the assets?
    \answerNA{NA.}
  \item Did you include any new assets in the supplemental material or as a URL?
    \answerNA{NA.}
  \item Did you discuss whether and how consent was obtained from people whose data you're using/curating?
    \answerYes{Yes.}
  \item Did you discuss whether the data you are using/curating contains personally identifiable information or offensive content?
    \answerYes{Yes.}
\item If you are curating or releasing new datasets, did you discuss how you intend to make your datasets FAIR (see \citet{fair})?
\answerNA{NA.}
\item If you are curating or releasing new datasets, did you create a Datasheet for the Dataset (see \citet{gebru2021datasheets})? 
\answerNA{NA.}
\end{enumerate}

\item Additionally, if you used crowdsourcing or conducted research with human subjects, \textbf{without compromising anonymity}...
\begin{enumerate}
  \item Did you include the full text of instructions given to participants and screenshots?
    \answerNA{NA.}
  \item Did you describe any potential participant risks, with mentions of Institutional Review Board (IRB) approvals?
    \answerNA{NA.}
  \item Did you include the estimated hourly wage paid to participants and the total amount spent on participant compensation?
    \answerNA{NA.}
   \item Did you discuss how data is stored, shared, and deidentified?
   \answerNA{NA.}
\end{enumerate}

\end{enumerate}

\appendix

\section{Appendix}

\subsection{Further Discussion on Potential Broader Impact and Ethical Considerations}
This study comes with potential negative societal impacts and opportunities for misuse. The study's reliance on English-centric contexts could inadvertently promote a homogenized view of communication, neglecting the rich diversity of emoji use across different cultures and languages. This could marginalize non-English speaking cultures or those that use emojis differently. Furthermore, the ability of LMMs to mimic human-like use of emojis could be exploited to create more convincing bots or fake accounts on social media. These could be used for spreading misinformation, manipulating public opinion, or engaging in deceptive marketing practices. To mitigate these issues, we have discussed the limitations of the English-centric contexts and only provide the aggregate results. 

\subsection{Additional Analysis of Study 2}
To ensure the robustness and generalizability of our findings, we extend our investigation by introducing 14 additional topics, distinct from those in the primary analysis. This approach allows us to examine if the observed outcomes are consistent across a broader spectrum of subjects. Data for these topics are collected using the TikTok API, employing the same collection and pre-processing methodology outlined for the primary analysis. The data span from February 15$^{th}$, 2024, to March 15$^{th}$, 2024, ensuring our results reflect the most current trends on TikTok.

In particular, our expanded dataset includes 14 additional topics: art, beauty, books, cooking, dance, DIY, fashion, fitness, kids, movies, music, plants, queer culture, and self-care. To collect the additional data, we use 23 specific hashtags~\cite{conrad2023complete} related to these topics: \#painting, \#arttok, \#makeuptutorial, \#girltips, \#beautyhacks, \#GRWM, \#CookBookTok, \#ThrillerTok, \#NonFicTok, \#RecipeTok, \#dance, \#dancer, \#diy, \#diytiktok, \#fashion, \#fashiontok, \#fittok, \#gymtok, \#funnykids, \#RomComTok, \#HorrorTok, \#musictok, \#houseplant, \#lgbtq, \#queer, \#selfcare. By using the method, we collect additional 2,109 videos.

\begin{figure}[ht]
    \centering
    \includegraphics[width=\linewidth]{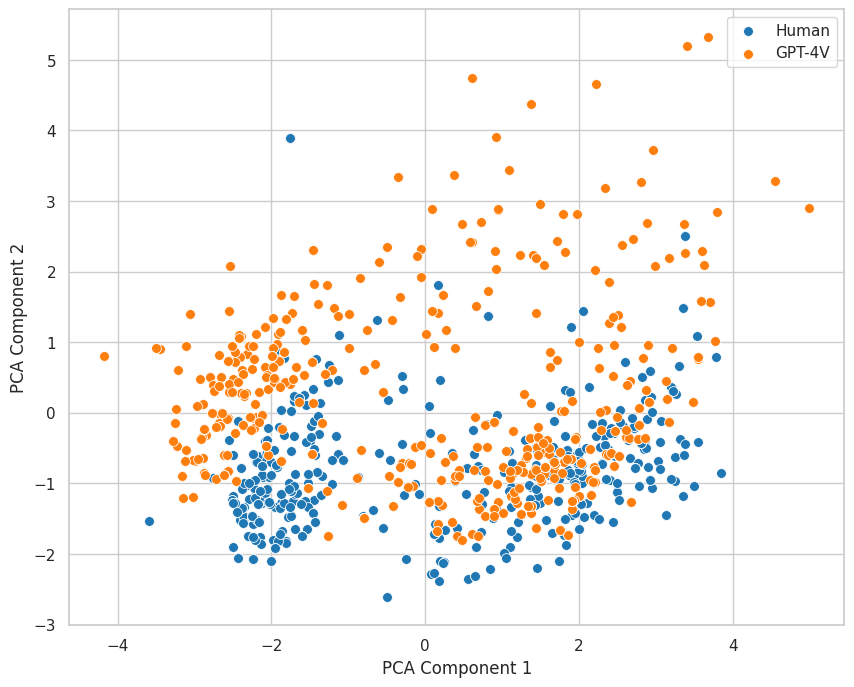}
    \caption{PCA visualization word embeddings of human-used and {\sc GPT-4V} generated emojis (additional data).}
    \label{fig: pca_appendix}
\end{figure}

We perform the PCA analysis on the extended dataset. The findings, depicted in Figure~\ref{fig: pca_appendix}, are {\it consistent} with the initial analysis. Additionally, Table~\ref{tab:emb_dist_appendix} details the mean and median distances between embeddings, along with their standard errors, offering a comparison to the results presented in Table~\ref{tab:emb_dist}. It is important to note that the standard errors tend to be larger in this analysis compared to the primary study. This variance can be attributed to the reduced number of videos per topic, a result of distributing the collection across a wider range of topics.

\begin{table}[ht]
\centering
\adjustbox{max width=\linewidth}{
\begin{tabular}{{l}c*{2}{c}r}
\toprule[1.1pt]
Hashtag   & Mean & Median & Standard Error \\
\midrule
GRWM & 0.2958 & 0.2955 & 0.0257 \\
HorrorTok  &  0.3102 & 0.3028   &   0.0325 \\
RecipeTok  & 0.2767  & 0.2683  & 0.0168 \\
ThrillerTok  &  0.2821  & 0.2511  & 0.0181  \\
arttok    &  0.2690   &  0.2814  &  0.0114 \\
beautyhacks  &   0.2946   &   0.2690  & 0.0169 \\
dance  &    0.2433 &   0.2546  &  0.0175 \\
dancer   &    0.3307  & 0.3281  &  0.0182 \\
diy    &     0.2971 &  0.2922 & 0.0229  \\
diytiktok   &  0.2961     &  0.2904  &  0.0132 \\
fashion   &  0.3110 &  0.2897 &  0.0246 \\
fashiontok & 0.2823   &  0.3003 &  0.0197 \\
fittok    &    0.2666   &  0.2569  &  0.0150 \\
funnykids   &   0.2671  &  0.2594 &  0.0122 \\
girltips   &  0.2929    &  0.2892 & 0.0191  \\
gymtok   &   0.3309     & 0.3363  &  0.0244 \\
houseplant  &  0.2376   & 0.2874 &  0.0222 \\
lgbtq          &   0.2989   & 0.2983  &  0.0170 \\
makeuptutorial  &  0.2907   &  0.2870 &  0.0143 \\
musictok   &    0.2711  &  0.2673 &  0.0172 \\
painting     &     0.2704   & 0.2819  &  0.0205 \\
queer   &     0.2963    & 0.2973  &  0.0267 \\
selfcare   &  0.2535    &   0.2436   &  0.0159  \\
\bottomrule[1.1pt]
\end{tabular}}
\caption{The mean and median cosine distances between the emojis used by humans and {\sc GPT-4V} (additional data).} 
\label{tab:emb_dist_appendix} 
\end{table}

\end{document}